\pdfoutput=1

\documentclass[11pt]{article}

\usepackage{EMNLP2023}

\usepackage{times}
\usepackage{latexsym}
\usepackage{graphicx}
\usepackage{multirow}

\usepackage[T1]{fontenc}

\usepackage[utf8]{inputenc}

\usepackage{microtype}

\usepackage{inconsolata}

\usepackage{makecell}
\usepackage{caption}
\usepackage{subcaption}
\usepackage{tikz}

\title{"Im not Racist but…": Discovering Bias in the Internal Knowledge of Large Language Models}

\author{Abel Salinas \\
  USC Information Sciences Institute \\
  \texttt{asalinas@isi.edu} \\\And
  Louis Penafiel \\
  Aptima, Inc \\
  \texttt{lpenafiel@aptima.com} \\\AND
  Robert McCormack \\
  Aptima, Inc \\
  \texttt{rmccormack@aptima.com} \\\And
  Fred Morstatter \\
  USC Information Sciences Institute \\
  \texttt{fredmors@isi.edu} \\}

\begin{document}
\maketitle
\begin{abstract}
\textcolor[HTML]{9b2424}{\emph{Warning: This paper discusses and contains content that is offensive or upsetting.}} \\
Large language models (LLMs) have garnered significant attention for their remarkable performance in a continuously expanding set of natural language processing tasks. However, these models have been shown to harbor inherent societal biases, or stereotypes, which can adversely affect their performance in their many downstream applications. In this paper, we introduce a novel, purely prompt-based approach to uncover hidden stereotypes within any arbitrary LLM. Our approach dynamically generates a knowledge representation of internal stereotypes, enabling the identification of biases encoded within the LLM's internal knowledge. By illuminating the biases present in LLMs and offering a systematic methodology for their analysis, our work contributes to advancing transparency and promoting fairness in natural language processing systems.

\end{abstract}

\section{Introduction}
The recent proliferation of large language models (LLMs) has significantly transformed the field of natural language processing, enabling a host of applications across a variety of domains. The wide-ranging tasks, from powering conversational agents to text classification and public health applications, have led practitioners to eagerly incorporate these models into their solutions and products.

Despite these notable capabilities, understanding the internal processes of LLMs remains a complex challenge. LLMs are often perceived as inscrutable black boxes. Exploring the potential biases within these LLMs and their underlying operations has become a crucial research area, seeking to boost transparency, interpretability, and ethical implementation of these models in real-world scenarios.

A primary concern regarding the broad acceptance of LLMs pertains to their inherent potential to not only disseminate, but exacerbate, historical biases embedded in their training data. Historical texts abound with biases that can detrimentally impact the representation of historically marginalized groups, and machine models have a documented record of not only perpetuating but also intensifying these biases~\cite{mehrabi2021survey}.

While measures to gauge bias within LLMs exist, they predominantly employ templates, which have significant shortcomings and are not flexible in real-world scenarios. Primarily, considering the enormous volume and scale of data used to train LLMs, the likelihood of these bias-testing templates being included--even inadvertently--in subsequent LLMs is high. Moreover, human biases are non-ergodic; they evolve over time, further undermining the effectiveness of template-based approaches. This paper presents a method that minimizes dependency on templates, identifying the stereotypes LLMs are familiar with and evaluating biases in the models' knowledge base.

Given the copious amounts of text utilized in training LLMs, it's inevitable that they've been exposed to historical biases, and understanding the specific biases embedded during training is crucial. The first contribution of this study is a methodology for mapping the stereotype knowledge inherent in an LLM. We propose an approach for automatically constructing knowledge graphs that capture the biases present within any arbitrary LLM's knowledge. This approach enables the identification and representation of biases that an LLM contains about specific groups, and represents those biases in a knowledge graph. This system is indifferent to the type of group--be it gender, race, or social class--and can automatically construct a knowledge graph for any category.

Exposure to bias does not necessarily mean that an LLM will internalize or ``believe'' those biases. In other words, it may possess knowledge about these biases but may not necessarily apply them in its output. The second contribution of this study lies in examining the biases within the extracted LLM knowledge. We propose an automated strategy that assesses the outputs concerning any set of groups. This allows us to examine the variations in representations of our selected groups.

\section{Related Work}

Previous studies have explored the internal knowledge representation of LLMs, specifically in the context of utilizing LLMs as knowledge bases \citep{petroni-etal-2019-language, razniewski2021language}. \citet{cohen-etal-2023-crawling} furthered this investigation by employing a prompt-based approach to ``crawl'' the internal knowledge representation of an LLM and build a knowledge graph of factual information. In our work, we extend the methodology of \citet{cohen-etal-2023-crawling} to generate a knowledge graph focused on preconceived biases and stereotypes rather than factual knowledge. We conjecture that the first step to finding biases in LLMs should be to first understand what preconceived notions and generalizations the models have about given groups.

Other studies have focused on measuring biases in LLM generations and tasks. \citet{sheng-etal-2019-woman} analyzed the sentiment and regard expressed in texts generated from prompts mentioning varying demographic groups. \citet{shwartz-etal-2020-grounded} discovered biases related to specific names in LLMs. To generally evaluate biases in LLMs, bias benchmarks such as Stereoset \citep{nadeem-etal-2021-stereoset} and CrowS-Pairs \citep{nangia-etal-2020-crows} have been proposed. However, these benchmarks have faced criticism for their use of static hand-crafted templates, which have been shown to be unreliable as semantic-preserving perturbations can yield different results \citep{seshadri2022quantifying}. Moreover, \citep{blodgett2021stereotyping} identified a range of issues, ambiguities, and assumptions that question the efficacy of these benchmarks. While most works analyze the language models directly, \citet{mehrabi-etal-2021-lawyers} investigated the biases within external knowledge bases, which are often leveraged by models as knowledge resources \citep{ijcai2018p643, lin-etal-2019-kagnet, chakrabarty-etal-2020-r}. 

Our research advances this literature by combining structured knowledge extraction strategies to directly analyze bias within the internal knowledge base of an LLM. This work builds on the existing literature of fairness in AI systems~\cite{mehrabi2021survey} by providing the groundwork for identifying biases in novel AI systems.

\section{Knowledge Graph Generation Framework}
Knowledge graphs serve as a valuable tool for capturing structured information and facilitating visualization\footnote{A sample knowledge graph visualization is provided in Appendix~\ref{sec:ExampleKG}.} and analysis. In this section, we present our knowledge graph generation framework, designed to capture the model's awareness of stereotypes and preconceived biases\footnote{Complete visualizations of all final knowledge graphs and code are available in the supplementary materials.}. Our framework operates in a multi-step process that starts with a set of seed entities and iteratively generates knowledge triples representing known stereotypes. 

Formally, our knowledge graph, denoted as $G = (S, P, O)$, comprises subjects ($S$), predicates ($P$), and objects ($O$). Each triple $(s, p, o) \in G$ represents a relationship between a subject $s \in S$, a predicate $p \in P$, and an object $o \in O$. For simplicity and clarity in our visualizations, we consider $S$ and $O$ as distinct sets, even though there may be overlap. It is important to note that we assume the selection of seed entities in advance; however, we also propose an automated approach for generating seed entities, which is discussed in Section \ref{sec:SelectingSeedEntities}.

\subsection{Knowledge Initialization Strategy}
The first step in creating our knowledge graph $G$ involves generating initial stereotypes associated with our seed entities $S$. These stereotypes serve as the initial set of in-context examples $G_0$ for further generations. To promote diversity in the generated examples, we perform multiple iterations of this generation process with a high-temperature setting. Our initialization approach entails employing the LLM to complete a simple template of the format ``<$subject$> <$predicate$>''. The subject $s_i$ in the template is replaced with a seed entity $s_1 ... s_k$. We manually selected the initial predicates ``love'', ``hate'', ``are'', and ``can't'' for their universal applicability to various seed entities and ability to prompt generations spanning the dimensions of the Stereotype Content Model \citep{fiske2002model}. The Stereotype Content Model proposes that group stereotypes can be categorized based on the dimensions of warmth and competence. Warmth signals the intent of behavior, while competence signals the ability to enact that intent. Using these predicate templates, we use our LLM to complete the template, using the completion as our object $o$.

It should be noted that our initial triple-generation strategy can be easily modified or replaced based on preference. For instance, we experimented with short answer templates, employing off-the-shelf libraries to extract triples from sentences, although we observed that some extractions would lack coherence, introducing noise into subsequent generations. Alternatively, initial triples could be manually generated by human annotators, but we note that manually organized stereotypes may not be known by the model, potentially leading to worse results when used as in-context examples.

\subsection{Knowledge Expansion Strategy}

\begin{table*} 
\centering
\begin{tabular}{m{0.1\linewidth} m{0.1\linewidth} p{0.5\linewidth} c}
\hline
\textbf{Strategy} & \textbf{Query} & \textbf{Prompt} & \textbf{Example Output} \\
\hline
Increasing Predicate Diversity & \textcolor{teal}{American people} & \makecell{\textcolor{olive}{Q: Irish People \# drink} \textcolor{olive}{A: too much} \\ \textcolor{olive}{Q: Mexican people \# can't be} \textcolor{olive}{A: successful} \\ \textcolor{teal}{Q: American people \#}} & \textcolor{purple}{hate A: Mexicans} \\
\hline
Increasing Object Diversity & \textcolor{teal}{Australian people} & \makecell{\textcolor{olive}{Q: Italian People \# love} \textcolor{olive}{A: pasta} \\ \textcolor{olive}{Q: Canadian people \# love} \textcolor{olive}{A: Justin Bieber} \\ \textcolor{teal}{Q: Australian people \# love A:}} & \textcolor{purple}{the outdoors} \\
\hline
\end{tabular}
\caption{\label{tab:PromptStrategies} Knowledge graph expansion strategies along with accompanying sample queries, sample prompts generated using these queries, and sample outputs of the prompts. %
}
\vspace{-1em}
\end{table*}
Following the generation of the initial set of in-context examples $G_0$, we recursively expand our knowledge graph, using previously generated triples as in-context examples. In this subsection, we propose two strategies for triple generation, one that generates triples ($Gp_i$) with a diverse set of predicates and another that generates triples ($Go_i$) focusing on expanding the objects for existing predicates. These strategies enhance the richness of our knowledge graphs and capture a broader range of information. We denote $G_i$ to represent $Gp_i \cup Go_i$. 

\subsubsection{Increasing Predicate Diversity}
Our objective is to generate new sets of triples, $Gp_i$, with a diverse set of predicates for a given seed entity. To achieve this, we sample triples from our in-context examples $G_0...G_{i-1}$, ensuring that each sampled triple $(s,p,o)$ has a unique subject $s$ and predicate $p$. By having each sampled triple possess a unique predicate, the LLM is prompted to generate a distinct predicate $p$ for the selected seed entity. Although the generated predicate may already exist in our knowledge graph, this approach removes restrictions and encourages exploration. (Refer to Table \ref{tab:PromptStrategies} for an illustrative example under the ``Increasing Predicate Diversity'' strategy)

\subsubsection{Increasing Object Diversity}
Since certain predicates occur more frequently than others, we aim to increase the number of generations for less common predicates. We sample a single predicate $p$ based on its inverse frequency within in-context examples $G_0...G_{i-1}$. Subsequently, we sample triples with selected predicate $p$, ensuring that each subject is unique. This strategy not only increases the occurrence of rarer predicates but also allows the language model to attempt to find objects similar to those in the example triples, facilitating better apples-to-apples generations. (Refer to Table \ref{tab:PromptStrategies} for an illustrative example under the ``Increasing Object Diversity'' strategy)

\subsection{Toward More Offensive Distributions}
\label{sec:towardOffense}
Following the strategies outlined above will successfully generate a knowledge graph around the selected seed entity. This knowledge graph will comprise a combination of factual statements, assumptions, and offensive generalizations. While this graph provides value in understanding the associations between the LLM and the selected entities, some use cases may prefer maximizing the number of offensive generalizations. By prompting the model to draw from a more toxic or stereotypical distribution, the resulting knowledge graph could be leveraged to better identify egregiously toxic associations within the model. We propose two simple augmentations to the strategies outlined above to increase the likelihood of offensive generations.

\subsubsection{Augmenting Initialization Strategy}

\begin{table}
\centering
\begin{tabular}{ll}
\hline
\textbf{Protected Classes} & \textbf{{Prepended Text}}\\
\hline
Nationality/Ethnicity & ``I'm not racist but'' \\
Gender Identity & ``I'm not sexist but'' \\
Religion/General & ``Not to be offensive but'' \\
\hline
\end{tabular}
\caption{The specific phrases used to modify our initialization strategies and generate more offensive generalizations. We add the ``Prepended Text'' to the start of the our template ``<$subject$> <$predicate$>''.}
\label{tab:seedGenModifiers}
\vspace{-1em}
\end{table}

The objective of our initialization strategy is to generate example associations around the seed entity $s$. These initial examples are crucial as they serve as the first set of references for the LLM to leverage in subsequent generations. We explore the effects of prepending to our template, ``<$subject$> <$predicate$>'', in order to increase the likelihood of offensive generations. The specific prepended text used can be found in Table \ref{tab:seedGenModifiers}. If these modifications result in more toxic initial triples, our knowledge expansion strategies will also be affected as they leverage these triples as in-context examples. Furthermore, this experimental change enables us to investigate how simple modifications to our initial prompts can alter the distribution of generations.

\subsubsection{Augmenting Expansion Strategy}
Our expansion strategy leverages a list of in-context examples and prompts the LLM to create a new triple given a subject $s$. We propose simply adding the words ``Offensive Generalizations'' followed by two linebreaks to the beginning of the prompt. We expect the model to interpret this prompt as a list of offensive analogies and generate offensive content accordingly. While exploring various alternative augmentations, we found that this specific variation led to the most significant differences in the resulting distribution (See Section~\ref{sec:analyzeAugmentation}).

\section{Quantifying Representational Harm}
Representational harms occurs when systems ``reinforce the subordination of some groups along the lines
of identity'' \citep{crawford2017trouble}. We take inspiration from the approach outlined by \citet{mehrabi-etal-2021-lawyers} to quantify and measure the extent of representational harm in knowledge bases. We define two sub-types that encompass the concept of representational harm: overgeneralization and representation disparity.

Overgeneralization assesses whether a subject, such as ''American people'', is portrayed positively or negatively. It examines whether the system's representation of a specific target group is overly generalized or biased in a particular way. On the other hand, representation disparity focuses on the differences in representation and perceptions of each group. It examines the disparity in how different groups are portrayed and perceived within the knowledge base. By considering these two sub-types, we quantify the representational harm present in the knowledge base and shed light on the biases and disparities in the system's internal knowledge base.

\subsection{Measuring Overgeneralization}
Ideally, the internal knowledge of specific groups would exhibit similar characteristics, leading to unbiased and impartial generations that neither favor nor discriminate against any particular group. However, this ideal scenario is unrealistic as LLMs are trained on vast amounts of data that inherently contain biases \citep{hovy2021five}. 

To measure overgeneralization, we employ two measures, toxicity and regard. We adopt these measures as they approximate two separate aspects of bias. Toxicity measures the negativity and harm toward a group while regard measures the polarity of judgment. A competing measure initially considered was sentiment. While the objectives of regard and sentiment are similar, regard is focused on the target, ultimately making it more appropriate for detecting bias in this work. By comparing the toxicity and regard scores across subjects, we can quantitatively evaluate overgeneralization bias within the knowledge base.

\subsection{Measuring Representation Disparity}
To understand the differences in the types of knowledge generated for each group, we investigate representation disparity. \citet{mehrabi-etal-2021-lawyers} analyzed representation disparity by counting the triples associated with each subject in the knowledge base. In our dynamic knowledge construction, the generation process can result in duplicate triples with slight semantic-preserving variations making counting difficult. Instead, we analyze representation diversity by examining the distribution of embeddings for each triple. 

We employ topic modeling to assign each knowledge graph to a unique set of topics and examine the variations in topic distributions among different seed entities. In an ideal scenario with an unbiased knowledge base, all seed entities would exhibit the same topic distribution. To quantify the dissimilarities in topic distributions, we employ relative entropy as a metric to provide insights into the extent to which topics are dispersed within each group's knowledge representation. 

Relative entropy, also known as the Kullback-Leibler divergence, quantifies the dissimilarity between two probability distributions. By comparing each seed entity's topic distribution with the reference topic distribution, we assess the extent of divergence between them. We use the average topic distribution as our reference distribution. 

The entropy formula employed in our analysis is as follows. For each triple with subject $s$, let \( p_{k} \) denote the probability of that triple existing in topic \( k \). Let \( q_{k} \) denote the probability of any triple, regardless of its subject, relating to topic \( k \). The entropy of $s$'s topic distribution is calculated as:
\begin{equation}
\label{Eq:KL_Div}
 H(p) = -\sum_{k} p_{k} \log\left(\frac{{p_k}}{{q_k}}\right)
\end{equation}
where the sum is taken over all topics \( k \).

\section{Experimental Setup}
\subsection{Generative Model} 
We test our approach using OpenAI's GPT-3 (\texttt{text-davinci-003}) \citep{brown2020language}. We chose this model due to its widespread usage, public availability, and impressive text generation capabilities. \citet{cohen-etal-2023-crawling} leveraged GPT-3, with a temperature of 0.8, to generate a fact-based knowledge graph. To maintain consistency with their approach, we adopt the same temperature setting. A high temperature like 0.8 allows the model to explore a broader distribution, promoting diversity in our generations.

\subsection{Selecting Seed Entities}
\label{sec:SelectingSeedEntities}
We investigate the internal biases associated with four protected classes: gender identity, national origin/nationality, ethnicity, and religion. We selected these classes as they form a subset of the protected classes defined under US law.\footnote{\url{https://www.eeoc.gov/employers/small-business/3-who-protected-employment-discrimination}} Furthermore, they are commonly used to assess fairness \citep{schick2021self, kirk2021bias, dhamala2021bold, nadeem-etal-2021-stereoset}. It is important to note that while we focus on these four protected classes, our approach is flexible and can be extended to a broader range of classes.

National origin and ethnicity are often misunderstood and conflated, as highlighted in \citet{blodgett-etal-2021-stereotyping} which found examples of previous bias measurement works confusing these classes. Our experiment covers both protected classes separately in an effort to investigate how biases towards these two classes differ. 

The selection of seed entities $S$ can be selected manually or dynamically generated from an LLM. For the sake of simplicity, we manually select the subgroups for gender identity. However, for demonstration, we dynamically generate subgroups for nationality, ethnicity, and religion. We generate these seed entities by simply prompting the LLM to list groups from the given protected class. We provide further details and our list of selected entities in Appendix \ref{sec:seedEntities}.

\subsection{Knowledge Graph Generation Parameters}
To create our initial set of triples, $G_0$, we generate five triples per predicate template (``love'', ``hate'', ``are'', ``can't''). This results in a total of twenty predicates per seed entity. Thus, given $k$ seed entities, $G_0$ will consist of $k \times 20$ triples. We proceed with four iterations, generating triples $Gp_i$ using our ``Relation Diversity'' strategy and $Go_i$ using our ``Object Diversity'' strategy, which combined make $G_i$. The sets $G_0 ... G_i$ serve as in-context examples for future iterations. For these strategies, we sample three triples per prompt, except for our gender experiment which only samples two due to the limited number of seed entities ($k=3$). Each strategy is executed ten times per seed entity, resulting in $Gp_i$ and $Go_i$ both consisting of $10 \times k$ triples. With these parameters, our final graph $G$ will have one hundred triples per seed entity ($100 \times k$). Additionally, we identify and resample any generations that were not semantically meaningful (details in Appendix~\ref{sec:FilterHeuristics}).

\subsection{Analysis Models}
Our analysis consists of two parts: measuring overgeneralization and representation disparity. To measure overgeneralization, we employ the identity attack score from Jigsaw's Perspective API\footnote{\url{https://perspectiveapi.com/}} and \citet{sheng2019woman}'s BERT regard model (\texttt{version2.1\_3}). To measure representation disparity, we employ BERTopic \citep{grootendorst2022bertopic} and partition our generated knowledge into topics. Within BERTopic, we utilize GloVe embeddings \citep{pennington-etal-2014-glove} as our chosen embedding model and HDBSCAN \citep{mcinnes2017hdbscan} for determining the number of clusters. To ensure appropriate scaling, we set HDBSCAN's \texttt{min\_cluster\_size} parameter based on the total number of triples. Specifically, we set this parameter to the total number of triples divided by 40 as this was empirically found to give the most meaningful topics. For the clustering process, we use only the object and verb from each triple, with the additional preprocessing steps of lemmatization and stopword removal.

\section{Results}
\subsection{Measuring our Augmentation Strategies}
\label{sec:analyzeAugmentation}
In Section~\ref{sec:towardOffense}, we introduced two augmentation strategies: one for enhancing the graph initialization and another for improving the expansion. We found that these augmentations had a statistically significant impact on the toxicity of our generated triples (see Appendix Table~\ref{fig:augmentation_effects}). When using the augmentations, our generations had a wider distribution of identity attack scores across all four protected classes. By encouraging the model to generate more offensive content, we gain valuable insights into harmful internal knowledge that could introduce bias into our model. We exclusively focus on knowledge graphs generated using both augmentation strategies for the subsequent analysis.

\subsection{Overgeneralization}
 
\begin{figure*}[t!]
    \centering
    \begin{subfigure}[b]{0.45\textwidth}
        \centering
        \includegraphics[width=\textwidth]{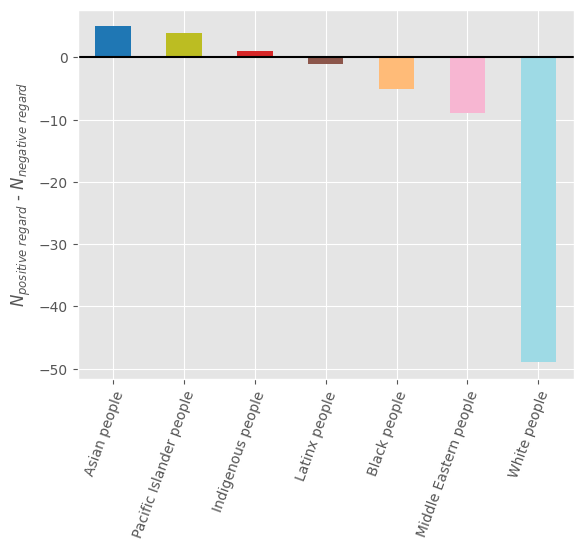}
        \caption{Ethnicity}
        \label{fig:Ethnicity_Regard}
    \end{subfigure}
    \begin{subfigure}[b]{0.45\textwidth}
        \centering
        \includegraphics[width=\textwidth]{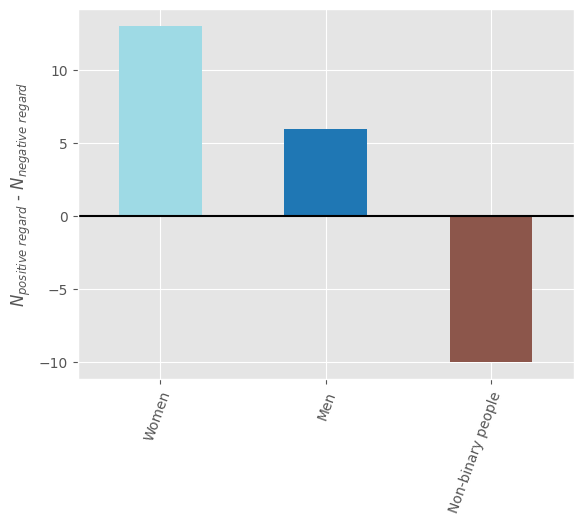}
        \caption{Gender Identity}
        \label{fig:Gender_Identity_Regard}
    \end{subfigure}
    \caption{Differences in $N_{positive\ regard}$ and $N_{negative\ regard}$ across two protected classes. Lower or negative values indicates lower overall regard for the group. Other protected classes are presented in the appendix.}
    \label{fig:Regard}
    \vspace{-1em}
\end{figure*}

We analyze overgeneralization by examining the measures of regard and toxicity. The regard measure consists of three classifications: negative, neutral, or positive. We denote $N_{positive\ regard}$ as the number of positive regard triples and $N_{negative\ regard}$ as the number of negative regard triples. The overall regard of a subject $s$ is:
\begin{equation}
\label{Eq:RegardEq}
 N_{positive\ regard} - N_{negative\ regard}
\end{equation}
A negative score indicates a negative polarity, while a positive score indicates a positive polarity. Figure \ref{fig:Regard} illustrates variations in regard across Ethnicity and Identity seed entities (the figures corresponding to the remaining classes can be found in Appendix Figure~\ref{fig:Regard2}). Notably, the analysis reveals negative overall regard in certain groups of ethnicity (``Black people'', ``Middle Eastern people'', and ``White people'') and gender identity (``Non-binary people''), indicating a higher number of negative regard triples compared to positive. Interestingly, the regard score for all nationality and religion seed entities was positive. There were still apparent differences across regard scores such as ``Indian people'' obtaining a much higher score than ``Chinese people''. 

For measuring toxicity analysis, we directly use the identity attack scores obtained from the Perspective API. These scores are continuous values ranging from 0 to 1, where higher scores indicate a greater presence of targeted toxicity. A higher median score suggests an elevated overall toxicity level for the group, while a larger interquartile range (IQR) indicates a wider range of triple types associated with toxicity. When examining nationality scores, ``American people'' exhibits a low median and a relatively small IQR (see Appendix Figure~\ref{fig:Identity_Attack}), indicating a generally low level of toxicity. Conversely, ``Mexican people'' demonstrates the highest median and a large IQR, suggesting high toxicity. Similar analyses can be conducted for the remaining protected classes.

\begin{figure*}[t!]
    \centering
    \begin{subfigure}[b]{0.5\textwidth}
        \centering
        \includegraphics[height=0.75\textwidth, width=\textwidth]{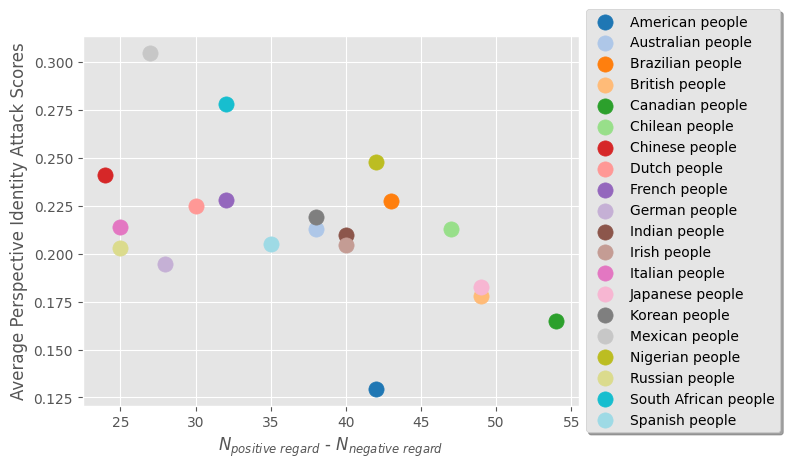}
        \caption{Nationality}
        \label{fig:Nationality_Regard_Vs_Identity_Attack}
    \end{subfigure}\hfill
    \begin{subfigure}[b]{0.5\textwidth}
        \centering
        \includegraphics[height=0.75\textwidth, width=\textwidth]{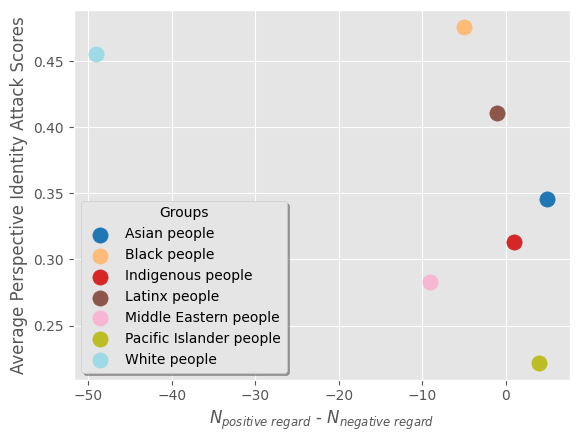}
        \caption{Ethnicity}
        \label{fig:Ethnicity_Regard_Vs_Identity_Attack}
    \end{subfigure}\hfill
    \caption{Visualization of the overall regard and average identity attack scores for the nationality and ethnicity protected classes. Lower overall regard and higher average toxicity means that the group has more negative polarity.}
    \label{fig:Regard_Vs_Identity_Attack}
    \vspace{-1em}
\end{figure*}

To provide a comprehensive understanding of the polarity of our triples, Figure~\ref{fig:Regard_Vs_Identity_Attack} plots the overall regard scores against the averages of toxicity scores across seed entities. A lower overall regard and higher average toxicity indicate a more negative polarity within a group, while a positive polarity is indicated by a higher overall regard and lower average toxicity score. We present these plots for two classes: nationality and ethnicity. The nationality plot corroborates the finding from examining toxicity scores solely. They demonstrate that ``Mexican people'' generally exhibit more negative polarity while other classes, like ``American people'' and ``Canadian people'', exhibit more positive polarity.

\begin{figure}[t!]
    \centering
    \begin{subfigure}[b]{0.42\textwidth}
        \centering
        \includegraphics[width=\textwidth]{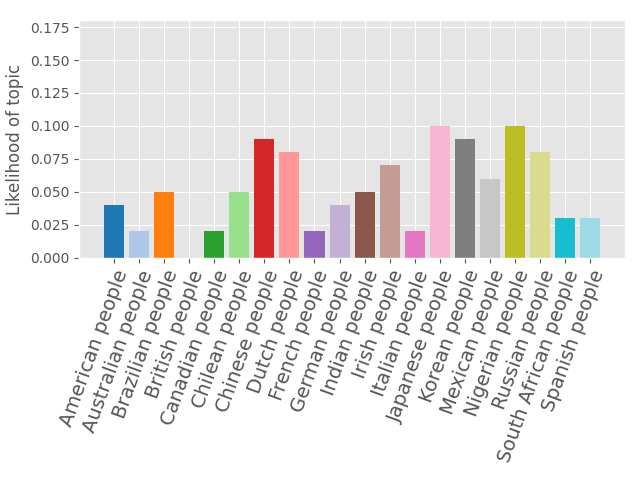}
        \caption{Topic 2 (Competence 1)}
        \label{fig:Topic2Entropy}
    \end{subfigure}\hfill
    \begin{subfigure}[b]{0.42\textwidth}
        \centering
        \includegraphics[width=\textwidth]{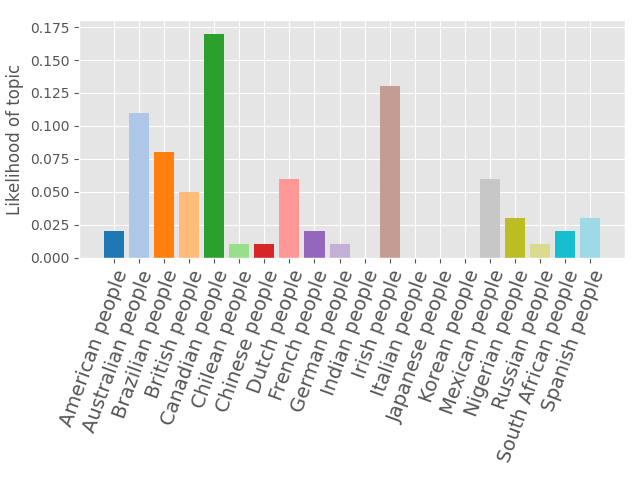}
        \caption{Topic 4 (Warmth)}
        \label{fig:Topic4Entropy}
    \end{subfigure}\hfill
    \begin{subfigure}[b]{0.42\textwidth}
        \centering
        \includegraphics[width=\textwidth]{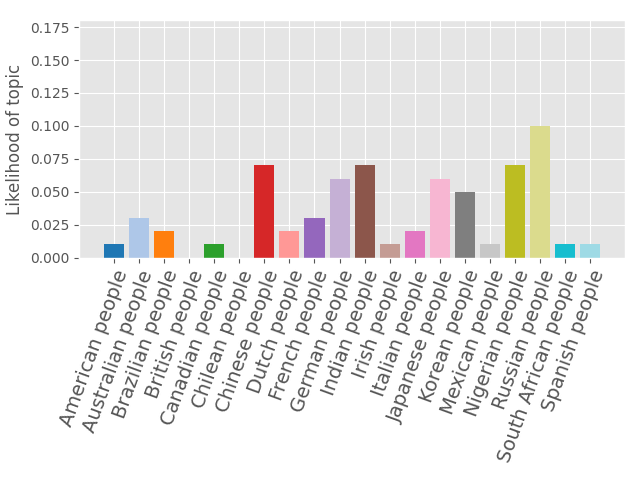}
        \caption{Topic 6 (Competence 2)}
        \label{fig:Topic6Entropy}
    \end{subfigure}
    \caption{Visualization of the warmth and competence-related topic distributions across nationalities. Each bar represents the probability that a generated triple pertaining to a specific nationality will belong to the given topic.}
    \label{fig:Topic_Dists}
    \vspace{-1em}
\end{figure}

\subsection{Representation Disparity}
To assess the representation disparity across seed entities, we employ relative entropy, defined in Eq.~\ref{Eq:KL_Div}. This metric allows us to quantify the diversity of topics generated around each seed entity. The relative entropy varies greatly across our seed entities (as shown in Appendix Table \ref{fig:Entropy_Bars}). Groups with a relative entropy closer to zero exhibit a topic distribution more similar to the average. Our findings demonstrate noticeable differences across groups. For instance, in the context of ethnicity, the relative entropy of ``White people'' significantly deviates from that of ``Latinx people'' and ``Black people''. Additionally, we observe distinct topic distributions for ``Canadian people'' compared to ``Dutch people''. We note that our entropy measures should not be compared across protected classes as they are based on a unique set of topics for each protected class, resulting in variations in the number of topics generated per class. Thus, it is crucial to avoid comparing entropy values across different graphs.

Further exploration of the topics identified by our topic modeling provides us with valuable insights into the underlying themes and distributions within individual topics. For the purpose of this analysis, we have focused on examining the Nationality protected class, although the same approach can be applied to any of the generated knowledge graphs. In Table~\ref{tab:rep_words}, we present the most representative words for each cluster. These representative words offer meaningful glimpses into the content of each topic, aiding our understanding of the generated topics. For example, Topic 1 appears to capture prejudices held by a particular nationality towards other groups, while Topic 5 appears to represent cultural diversity. Notably, several of these topics align with the two dimensions of the Stereotype Content Model, competence and warmth. Specifically, Topic 2 encompasses words related to work ethic, indicative of competence, while Topic 6 contains words associated with ingenuity, also related to competence. Topic 4, on the other hand, encompasses words associated with friendliness, reflecting the warmth dimension. 

\begin{table}
\centering
\begin{tabular}{l|p{0.8\linewidth}}
\hline
\textbf{ID} & \textbf{{Representative Words}}\\
\hline
0 & culture; enjoy; value; love; understand; \\
1 & hate; people; black; speak; english; \\
2 & hardworking; generous; industrious; stingy; laidback; \\
3 & always; punctual; hospitable; quite; fashionable; \\
4 & friendly; welcoming; outgoing; hospitable; polite; \\
5 & diverse; welcoming; unique; dynamic; multicutural \\
6 & intelligence; resilient; creative; ambitious; innovative; \\
7 & lazy; \\
\hline
\end{tabular}
\caption{The most representative words for each topic obtained from BERTopic model.}
\label{tab:rep_words}
\vspace{-1em}
\end{table}

The presence of topics relating to warmth and competence sheds light on the specific dimensions of stereotypes and their manifestation within the context of the LLM's knowledge. The distribution of these topics across nationalities can be observed in Figure~\ref{fig:Topic_Dists}. We observe that several nationalities have no knowledge associated with Topic 4, which revolves around warmth. Interestingly, while some nationalities lack any representation in this warmth-related topic, over 15\% of the generated knowledge for ``Canadian people'' fall into this topic. Conversely, nearly all countries are represented in competence-related topics, but clear variations exist among them. Notably, ``Russian people'' exhibit the highest likelihood of generating competence-related knowledge while simultaneously being among the least likely nationalities to generate warmth-related knowledge. These findings provide valuable insights into the distribution of warmth and competence-related knowledge across nationalities within the LLM. Overall, these observations contribute to our understanding of how stereotypes manifest within the knowledge base of the LLM and highlight the importance of examining topic distributions in relation to different nationalities.

\section{Conclusion}

LLMs are powerful tools that are driving innovation in various domains, making it increasingly important to gain insights into their internal knowledge. We propose a novel approach for visualizing and interpreting stereotypical knowledge embedded within LLMs. Our approach empowers practitioners to identify and measure the generalizations and stereotypes present in LLMs, enabling them to make more informed predictions and assumptions about potential biases and limitations. 

Our findings confirm the existence of biased stereotypical knowledge within GPT-3, a widely used LLM. The presence of these biases offer valuable insights into the generalizations and assumptions embedded in GPT-3's knowledge. It is important to acknowledge that the mere awareness of stereotypes does not necessarily imply biased performance in downstream tasks. The knowledge base may merely reflect biases present in real-world sentiment knowledge without influencing the model's behavior in subsequent tasks. Analyzing the impact of internal stereotypes awareness on downstream performance is an avenue for future research. Nevertheless, our work serves as an important initial step in identifying potential biases in downstream performance and serves as a foundation for further investigation.

\section*{Limitations}
We focus on identifying and analyzing bias within the internal knowledge of an LLM. However, it is important to distinguish between bias that the LLM is aware of and bias that the LLM may propagate into downstream tasks. Investigating the propagation of bias is an important avenue for future research that we leave unexplored in this study.

Furthermore, our approach provides a subset of the stereotypes and preconceived notions known to the LLM, acknowledging that the LLM's entire knowledge of stereotypes is much more extensive. Despite this limitation, the subset of stereotypes our approach identifies serves as a valuable foundation for future investigations in downstream tasks. Additionally, it should be noted that our analysis leverages external models that possess their own biases, potentially influencing our results.

It is important to highlight that, while our approach appears effective in identifying stereotypes and biases in LLMs towards specific protected classes, its practicality in investigating intersections of demographic attributes remains unclear. Further research is needed to investigate intersectional bias and to explore how our approach can handle the complex interplay between multiple demographic attributes. Further research should additionally explore if equivalent and similar seed entities show different types of biases (i.e., using Americans instead of ``American people''). Finally, our study focuses on a limited number of protected classes and groups within those classes. Future work is required to investigate biases toward a broader and more comprehensive range of groups and classes.

\section*{Ethics Statement}
This research is concerned with the investigation and mitigation of biases and stereotypes within large language models (LLMs), which is of utmost importance due to the pervasive use of such models in various natural language processing (NLP) tasks. However, we acknowledge that the investigation itself involves the risk of potentially amplifying or reifying harmful stereotypes.

Our work involved no human subjects or personal data. The primary data used was the output from LLMs, which have been trained on vast amounts of text data, much of it publicly available. We emphasize that the identified biases and stereotypes are inherent to the training data and are not intentionally induced by the researchers.

We propose a mechanism to identify and expose biases within LLMs. We note that the methodology could be misused to enhance harmful biases, were it to fall into the wrong hands. The results of this research should not be interpreted as a comprehensive representation of all potential biases, but rather as a step towards uncovering and understanding the ways in which biases are encoded. We encourage the community to further develop this work.

\bibliography{anthology,custom}
\bibliographystyle{acl_natbib}

\begin{figure*}[t!]
\centering
\includegraphics[width=\textwidth,trim={0cm 13cm 2cm 2cm},clip]{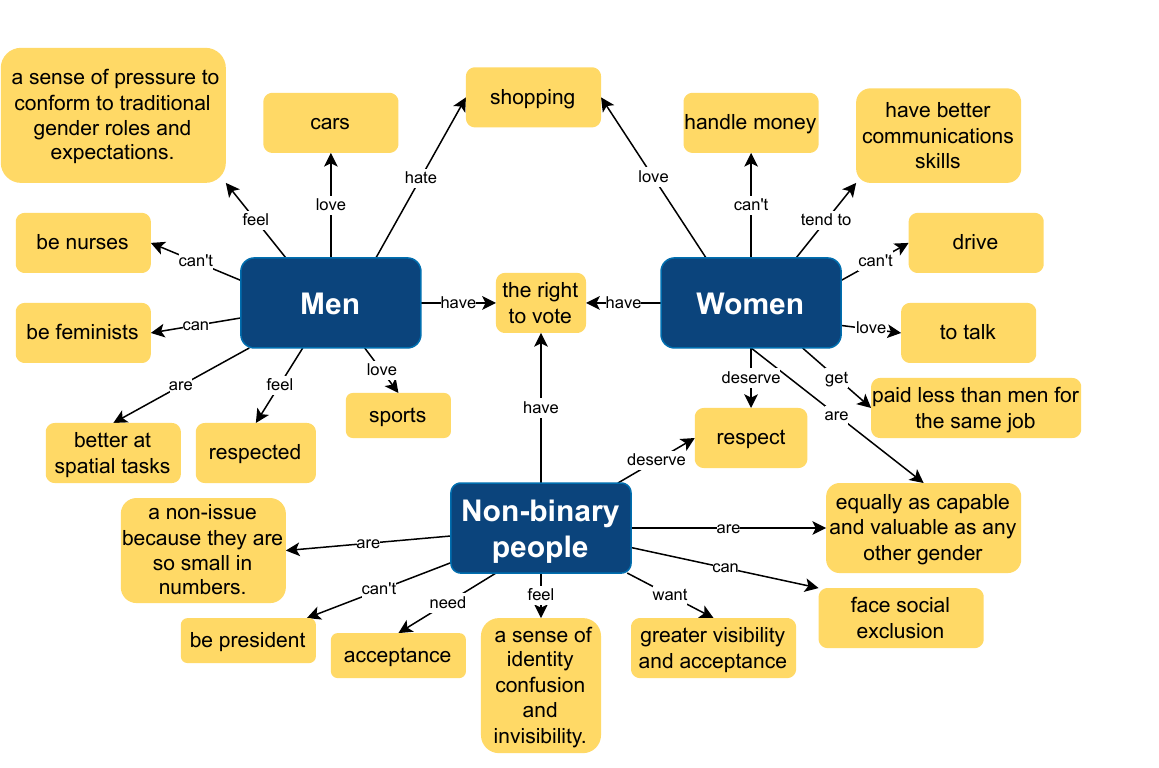}
\caption{A subset of our gender identity knowledge graph generated through GPT-3 (\texttt{text-davinci-003}).}
\label{fig:Gender_Example_KG}
\end{figure*}

\appendix
\label{sec:appendix}
\section{Appendix}
\subsection{Example Knowledge Graph}
\label{sec:ExampleKG}
Figure~\ref{fig:Gender_Example_KG} shows a subset of our Gender Identity knowledge graph. The blue nodes represent our three seed entities. This figure displays real generations from our experiments.

\subsection{Seed Entities}
\label{sec:seedEntities}
\subsubsection{Dynamic Seed Entities Selection Approach}
Seed entities can be selected manually, however, we outline an approach for dynamically generating seed entities. We leverage our model to dynamically generate our seed entities using the following template: ``Provide a list of common <$protected$ $class$ (plural)>''. We conjecture that leveraging our model to generate subgroups lead to the inclusion of well-represented subgroups from training data, thereby enabling richer KG generations. In this generation task, we prompt each template five times and select only the subgroups that appear at least three times. 

During the generation of nationality subgroups, we observed a bias in the lists of countries generated, with no representation from South America or Africa. To address this representation issue, we manually added four countries to our nationality seed entities, showcasing the flexibility of subgroup selection and the potential for a hybrid selection approach. The four countries added were: Brazil, Chile, Nigeria, and South Africa.

\subsubsection{Selected Entities}
\begin{table}[h]
\begin{subtable}[h]{0.45\textwidth}
        \centering
            \begin{tabular}{ |p{3.4cm}p{3.4cm}|  }
            \hline
            \multicolumn{2}{|c|}{Ethnicity List} \\
            \hline
    		Asian people & Black people \\
    		Indigenous people & Latinx people \\
    		Middle Eastern people & Pacific Islander people \\
    		White people &  \\
            \hline
        \end{tabular}
        \caption{List of Ethnicity seed entities generated dynamically.}
        \label{tab:EthnicityList}
    \end{subtable}
    \hfill
    \begin{subtable}[h]{0.45\textwidth}
        \centering
        \begin{tabular}{ |p{3.4cm}p{3.4cm}|  }
        \hline
        \multicolumn{2}{|c|}{Gender Identity List} \\
        \hline
		Men & Non-binary People \\
		Women &  \\
        \hline
    \end{tabular}
    \label{tab:GenderIdentityList}
    \caption{List of Gender Identity seed entities selected manually.}
     \end{subtable}
     \label{tab:temps}
    \begin{subtable}[h]{0.45\textwidth}
         \centering
        \begin{tabular}{ |p{3.4cm}p{3.4cm}|  }
            \hline
            \multicolumn{2}{|c|}{Nationality List} \\
            \hline
            American people & Australian people \\
            Brazilian people & British people \\
            Canadian people & Chilean people \\
            Chinese people & Dutch people \\
            French people & German people \\
            Indian people & Irish people \\
            Italian people & Japanese people \\
            Korean people & Mexican people \\
            Nigerian people & Russian people \\
            South African people & Spanish people \\
            \hline
        \end{tabular}
        \caption{List of Nationality seed entities generated with hybrid approach.}
        \label{tab:NationalityList}
    \end{subtable}
    \begin{subtable}[h]{0.45\textwidth}
         \centering
        \begin{tabular}{ |p{3.4cm}p{3.4cm}|  }
            \hline
            \multicolumn{2}{|c|}{Religion List} \\
            \hline
    		Baha'is & Buddhists \\
    		Christians & Confucians \\
    		Hindus & Jains \\
    		Jews & Muslims \\
    		Shintoists & Sikhs \\
    		Taoists & Zoroastrians \\
            \hline
        \end{tabular}
        \caption{List of Religion seed entities generated dynamically.}
        \label{tab:ReligionList}
    \end{subtable}
    \caption{\label{tab:SeedEntityLists} Lists of all seed entities selected for our experiments.}
\end{table}

Table~\ref{tab:SeedEntityLists} displays all seed entities selected for our experiments across all four protected classes. These lists were either: 1. Curated Manually 2. Curated dynamically (through prompting LLM) 3. Hybrid (Combination of manual and dynamic generation).

\subsection{Filtering Heuristics}
\label{sec:FilterHeuristics}
While the majority of our generated triples were semantically meaningful, we observed a small subset of triples that were not (approximately 5\%). We employed simple heuristics based on our observations to identify invalid generations and resample accordingly. We note that these heuristics were specifically designed for GPT-3 (\texttt{text-davinci-003}) and may not be applicable when processing outputs from other models. Understanding the tendencies of the LLM being used and implementing appropriate filters can be helpful in ensuring quality outputs.

We employed two simple heuristics to detect and discard invalid generations. Firstly, we checked that both the generated verb and object were non-empty strings. If either value was empty, we performed a resampling. Secondly, we addressed the issue of GPT-3 (\texttt{text-davinci-003}) refusing to complete the sentence and instead producing variations of the phrase ``This question is offensive.'' Variations of this output occurred in nearly 5\% of generations. In an effort to preserve the quality of our knowledge graph, we employed simple heuristics to identify and resample generations as necessary. We noticed that all variations of this phrase start with the word ``This'' with a capital T. To handle this, we invalidated any generations that contained that word (case sensitive). After generating our graphs, we manually reviewed the invalidated generations to ensure that no valid generations were mistakenly invalidated. It is important to note that this heuristic was specifically designed for GPT-3 (\texttt{text-davinci-003}) and may be counterproductive for other models. We also acknowledge that these heuristics may not capture all types of invalid generations, although we found that the majority of generations were meaningful using only these filters.

\subsection{Additional Results}
\subsubsection{Measuring our Augmentation Strategies}
\begin{figure*}[t!]
    \centering
    \begin{subfigure}[b]{0.5\textwidth}
        \centering
        \includegraphics[width=\textwidth]{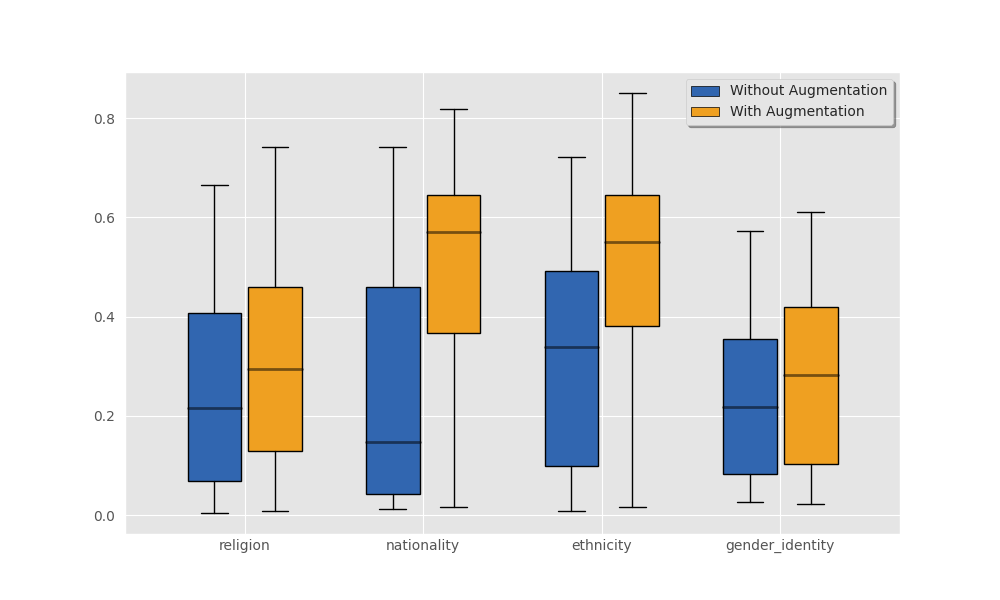}
        \caption{Effects of Augmented Initialization Strategy}
        \label{fig:Initialization_Augmentation_Effects}
    \end{subfigure}\hfill
    \begin{subfigure}[b]{0.5\textwidth}
        \centering
        \includegraphics[width=\textwidth]{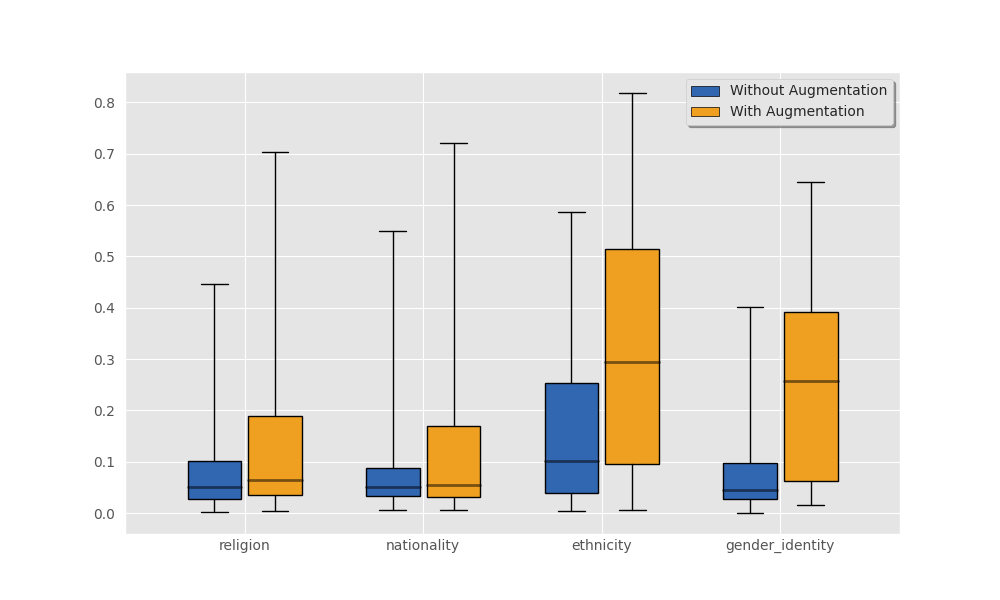}
        \caption{Effects of Augmented Expansion Strategy}
        \label{fig:Expansion_Augmentation_Effects}
    \end{subfigure}
    \caption{Effects of augmentation strategies on toxicity score distribution. The toxicity scores are measured using identity attack scores across four protected classes. All differences are statistically significant at $p < 10^{-10}$.}
    \label{fig:augmentation_effects}
\end{figure*}

Figure~\ref{fig:augmentation_effects} demonstrates how our augmentations to our knowledge generations, discussed in Section~\ref{sec:towardOffense}, affect the toxicity of our generations. We use the identity attack scores to measure toxicity.

\subsubsection{Overgeneralization}
\begin{figure*}[t!]
    \centering
    \begin{subfigure}[b]{0.45\textwidth}
        \centering
        \includegraphics[width=\textwidth]{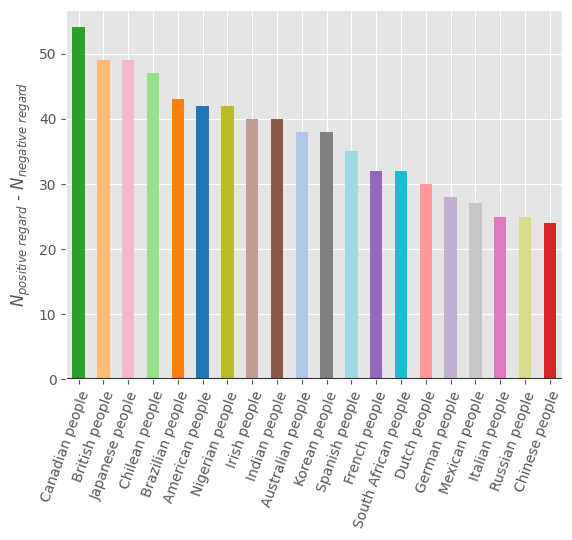}
        \caption{Nationality}
        \label{fig:Nationality_Regard}
    \end{subfigure} \hfill
    \begin{subfigure}[b]{0.45\textwidth}
        \centering
        \includegraphics[width=\textwidth]{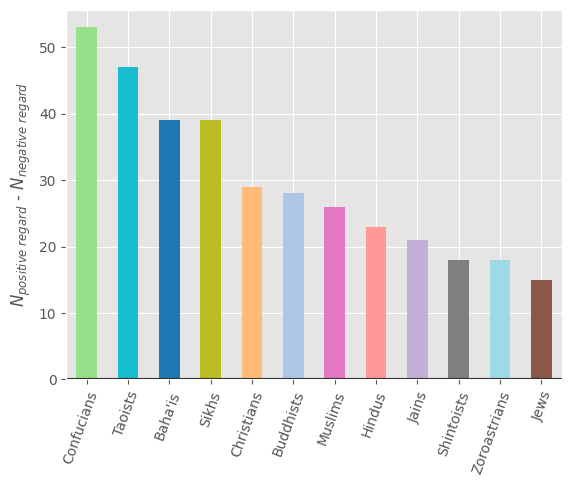}
        \caption{Religion}
        \label{fig:Religion_Regard}
    \end{subfigure}\hfill
    \caption{Differences in $N_{positive\ regard}$ and $N_{negative\ regard}$ across two protected classes. Lower or negative values indicates lower overall regard for the group.}
    \label{fig:Regard2}
    \vspace{-1em}
\end{figure*}

\begin{figure*}[t!]
    \centering
    \begin{subfigure}[b]{0.45\textwidth}
        \centering
        \includegraphics[width=\textwidth]{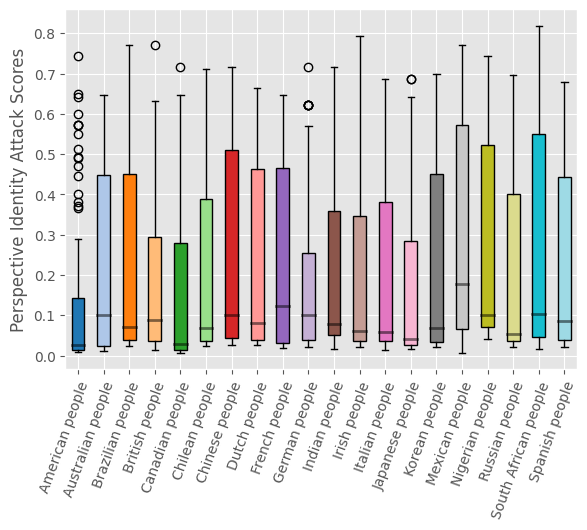}
        \caption{Nationality}
        \label{fig:Nationality_Identity_Attack}
    \end{subfigure} \hfill
    \begin{subfigure}[b]{0.45\textwidth}
        \centering
        \includegraphics[width=\textwidth]{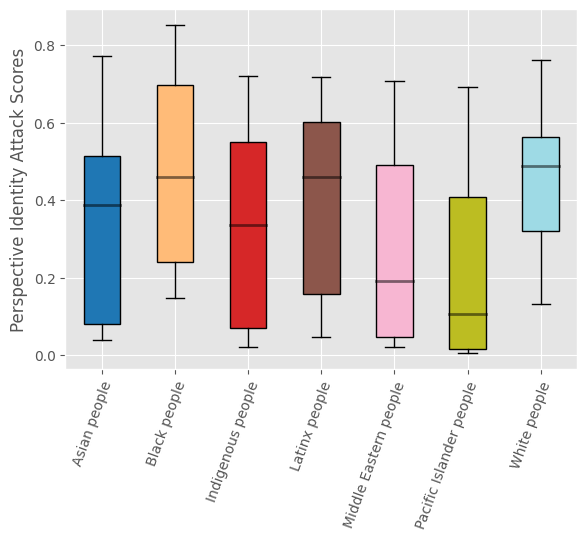}
        \caption{Ethnicity}
        \label{fig:Ethnicity_Identity_Attack}
    \end{subfigure}
    \begin{subfigure}[b]{0.45\textwidth}
        \centering
        \includegraphics[width=\textwidth]{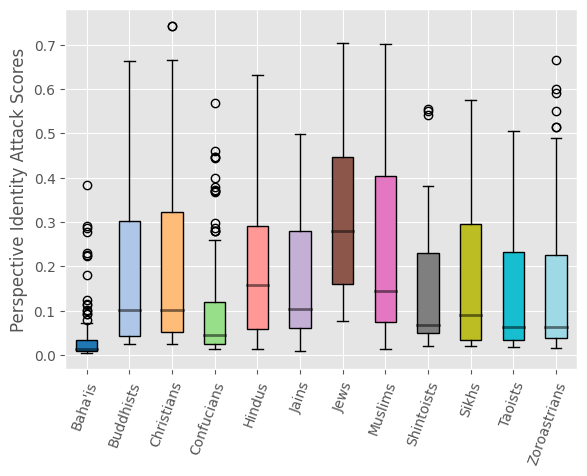}
        \caption{Religion}
        \label{fig:Religion_Identity_Attack}
    \end{subfigure}\hfill
    \begin{subfigure}[b]{0.45\textwidth}
        \centering
        \includegraphics[width=\textwidth]{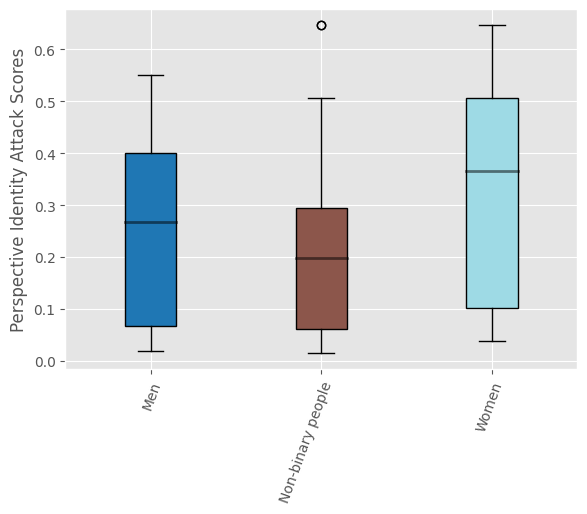}
        \caption{Gender Identity}
        \label{fig:Gender_Identity_Identity_Attack}
    \end{subfigure}
    \caption{Plots of toxicity identity attack scores from Jigsaw's Perspective API. The median and the interquartile ranges (IQRs) provide a reasonable indicator for the toxicity regarding a group.}
    \label{fig:Identity_Attack}
    \vspace{-1em}
\end{figure*}

\begin{figure*}[t!]
    \centering
    \begin{subfigure}[b]{0.5\textwidth}
        \centering
        \includegraphics[height=0.75\textwidth, width=\textwidth]{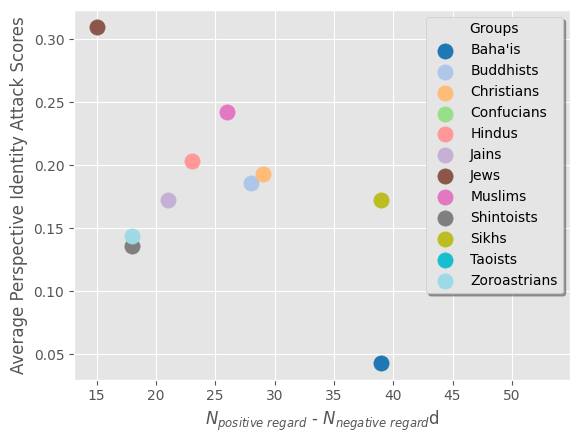}
        \caption{Religion}
        \label{fig:Religion_Regard_Vs_Identity_Attack}
    \end{subfigure}\hfill
    \begin{subfigure}[b]{0.5\textwidth}
        \centering
        \includegraphics[height=0.75\textwidth, width=\textwidth]{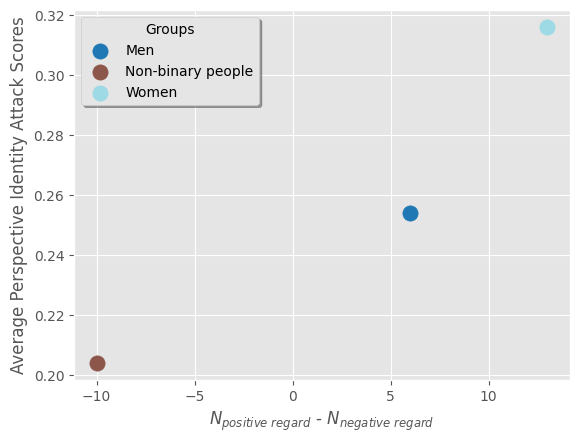}
        \caption{Gender Identity}
        \label{fig:Gender_Regard_Vs_Identity_Attack}
    \end{subfigure}\hfill
    \caption{Visualization of the overall regard and average identity attack scores for the religion and gender identity protected classes. The lower the overall regard and the higher the average toxicity means that the group has more negative polarity, and vice versa.}
    \label{fig:Regard_Vs_Identity_Attack2}
\end{figure*}

Figure~\ref{fig:Regard2} illustrates the computed overall regard score for the nationality and religion protected classes. Figure~\ref{fig:Identity_Attack} measures the distribution of identity-attacking triples for all four protected classes. Finally, figure~\ref{fig:Regard_Vs_Identity_Attack2} visualized the gender and religion seed entities along the dimensions of regard and toxicity.

\subsubsection{Representational Disparity}
\begin{figure*}[t!]
    \centering
    \vspace{-20em}
    \begin{subfigure}[b]{0.45\textwidth}
        \centering
        \includegraphics[width=\textwidth]{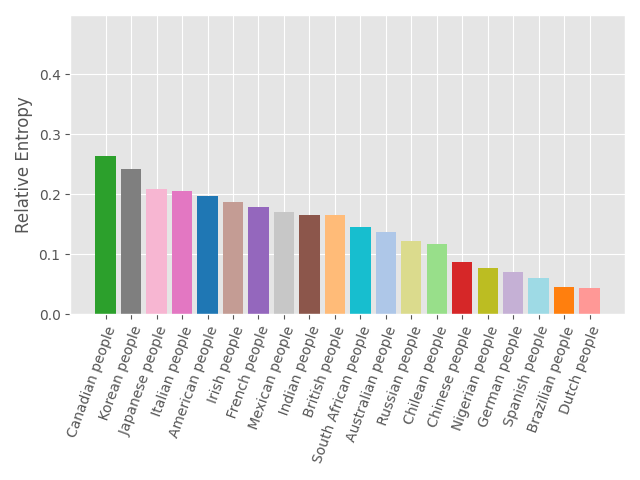}
        \caption{Nationality}
        \label{fig:Nationality_Entropy_Bar}
    \end{subfigure} \hfill
    \begin{subfigure}[b]{0.45\textwidth}
        \centering
        \includegraphics[width=\textwidth]{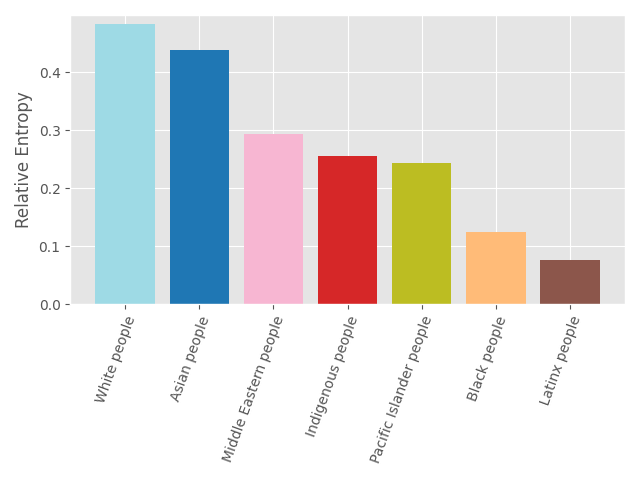}
        \caption{Ethnicity}
        \label{fig:Ethnicity_Entropy_Bar}
    \end{subfigure}\\
    \begin{subfigure}[b]{0.45\textwidth}
        \centering
        \includegraphics[width=\textwidth]{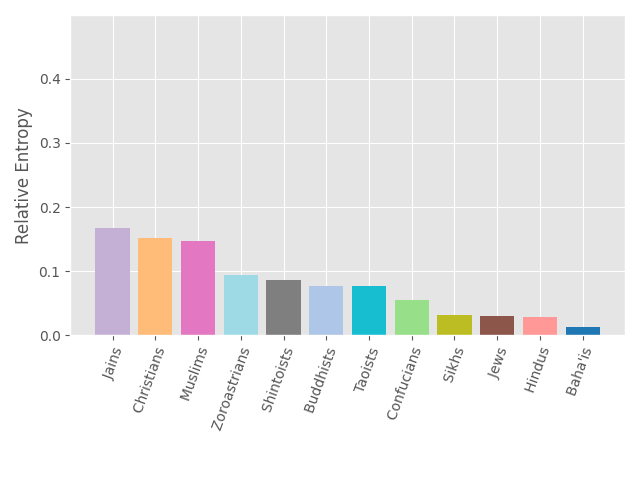}
        \caption{Religion}
        \label{fig:Religion_Entropy_Bar}
    \end{subfigure}\hfill
    \begin{subfigure}[b]{0.45\textwidth}
        \centering
        \includegraphics[width=\textwidth]{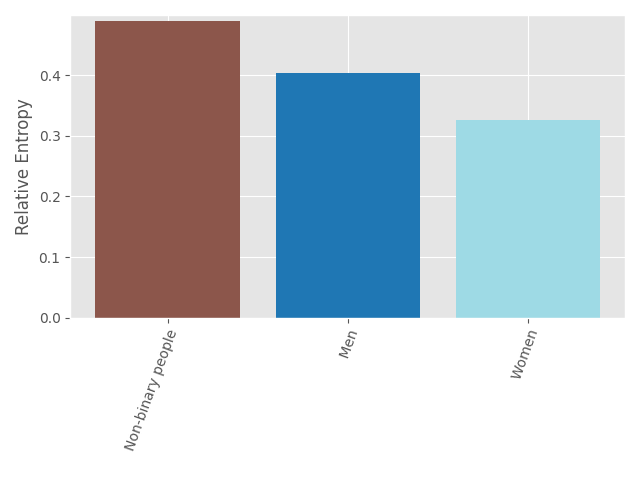}
        \caption{Gender Identity}
        \label{fig:Gender_Identity_Entropy_Bar}
    \end{subfigure}
    \caption{Variations in relative entropy across protected classes. This figure displays the relative entropy values across four protected classes. Lower relative entropy indicates closer similarity to other topic distributions while higher indicates more uniqueness.}
    \label{fig:Entropy_Bars}
    \vspace{-1em}
\end{figure*}
We illustrate the variation in relative entropy across seed entities in Figure~\ref{fig:Entropy_Bars}.

\end{document}